\documentclass[letterpaper]{article} 
\usepackage{aaai24}  
\usepackage{times}  
\usepackage{helvet}  
\usepackage{courier}  
\usepackage[hyphens]{url}  
\usepackage{graphicx} 
\usepackage{natbib}  
\usepackage{caption} 
\usepackage{algorithm}
\usepackage{algorithmic}

\usepackage{newfloat}
\usepackage{listings}
\DeclareCaptionStyle{ruled}{labelfont=normalfont,labelsep=colon,strut=off} 
\floatstyle{ruled}
\newfloat{listing}{tb}{lst}{}
\floatname{listing}{Listing}

\usepackage{enumitem}
\usepackage{color} 
\usepackage{graphicx}
\usepackage{booktabs}
\usepackage{subcaption}
\usepackage{todonotes}
\usepackage{array}
\usepackage{amssymb, amsmath}
\usepackage{multicol}
\usepackage{multirow}
\usepackage{threeparttable}
\usepackage{algorithm}
\usepackage{algorithmic}
\usepackage{microtype}
\usepackage{listings}

\definecolor{mypurple1}{RGB}{143, 94, 255}
\definecolor{myblue1}{RGB}{59, 76, 206}
\definecolor{myred1}{RGB}{190, 90, 75}
\definecolor{myyellow1}{RGB}{255, 192, 0}
\definecolor{mygreen1}{RGB}{160, 194, 128}
\nocopyright

\title{JoTR: A Joint Transformer and Reinforcement Learning Framework for Dialog Policy Learning}
\author{
    Wai-Chung Kwan\equalcontrib \textsuperscript{\rm 1},
    Huimin Wang\equalcontrib \textsuperscript{\rm 2},
    Hongru Wang \textsuperscript{\rm 1},
    Zezhong Wang \textsuperscript{\rm 1},
    Xian Wu \textsuperscript{\rm 2}, \\
    Yefeng Zheng \textsuperscript{\rm 2},
    Kam-Fai Wong \textsuperscript{\rm 1}
}
\affiliations{
    \textsuperscript{\rm 1} The Chinese University of Hong Kong\\
     \textsuperscript{\rm 2} Jarvis Lab, Tencent \\

}

\usepackage{bibentry}

\begin{document}

\maketitle

\begin{abstract}
Dialogue policy learning (DPL) is a crucial component of dialogue modelling. Its primary role is to determine the appropriate abstract response, commonly referred to as the “dialogue action”. Traditional DPL methodologies have treated this as a sequential decision problem, using pre-defined action candidates extracted from a corpus. However, these incomplete candidates can significantly limit the diversity of responses and pose challenges when dealing with edge cases, which are scenarios that occur only at extreme operating parameters. To address these limitations, we introduce a novel framework, JoTR. This framework is unique as it leverages a text-to-text Transformer-based model to generate flexible dialogue actions. Unlike traditional methods, JoTR formulates a word-level policy that allows for a more dynamic and adaptable dialogue action generation, without the need for any action templates. This setting enhances the diversity of responses and improves the system's ability to handle edge cases effectively. In addition, JoTR employs reinforcement learning with a reward-shaping mechanism to efficiently finetune the word-level dialogue policy, which allows the model to learn from its interactions, improving its performance over time. We conducted an extensive evaluation of JoTR to assess its effectiveness. Our extensive evaluation shows that JoTR achieves state-of-the-art performance on two benchmark dialogue modelling tasks, as assessed by both user simulators and human evaluators \footnote{Our code, models and other related resources are publicly available at https://github.com/KwanWaiChung/JoTR.}
\end{abstract}

\section{Introduction}
\begin{figure}
    \centering
    \includegraphics[width=0.45\textwidth]{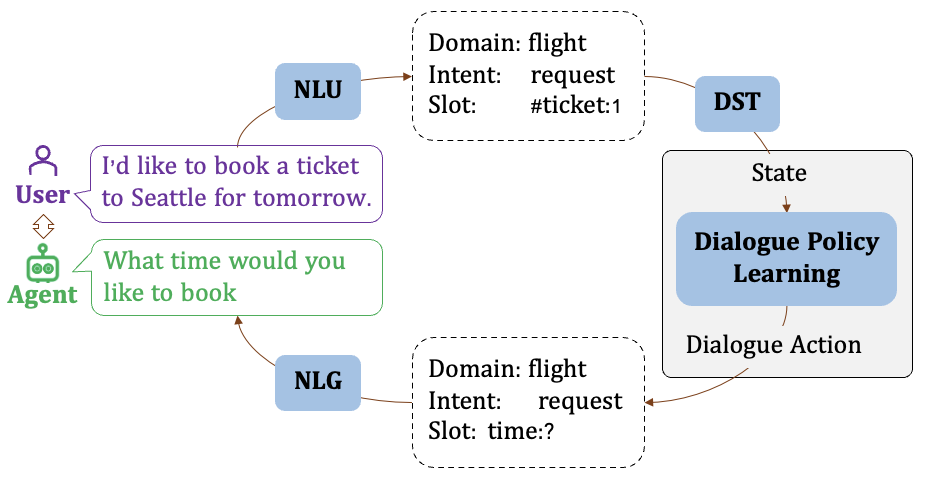}
    \caption{This illustrates dialogue policy learning in pipeline dialogue system, where NLU/NLG stands for Natural Language Understanding/Generation, and DST refers to Dialogue State Tracking.}
    \label{fig1:intro}
    \vspace{-2mm}
\end{figure}

Dialogue policy learning (DPL) is a fundamental component of pipeline-based dialogue systems. As illustrated in Figure \ref{fig1:intro}, its primary role is to select the appropriate dialogue action to manage the conversation flow, essentially acting as the decision-making core of the system. Previous research in this field has predominantly focused on treating DPL as a sequential decision problem. The optimization of the policy is typically achieved using reinforcement learning (RL), a machine learning approach that learns from its interactions with the environment, based on predefined dialogue action candidates \cite{lipton2016efficient, li2017end, peng2018adversarial,takanobu-etal-2019-guided, wang2020learning, li2020guided, kwan2023survey}. A dialogue action usually comprises one or more combinations of domain name, intent type, and slot name, collectively referred to as an “atomic action” \cite{li2020rethinking}. Traditional DPL approaches filter action candidates from a corpus, placing emphasis on more frequent actions to enhance the efficiency of the RL agent. However, this approach has its limitations. It may underperform in less common situations and can restrict response diversity and naturalness due to the rigid templates of allowable actions \cite{wang-wong-2021-collaborative}. This limitation stems from the fact that policy optimization is based on incomplete action candidates, which include only a subset of possible atomic actions or their combinations. To address this issue, \citet{li2020rethinking} proposed using a GRU-based decoder to predict atomic actions sequentially in a single turn. This approach allows the DPL model to learn the action structure from data, resulting in more flexible and powerful system responses. However, this method has a significant drawback: the dialogue state and atomic action space grow exponentially with the number of domains, intents, or slots \cite{lee-etal-2019-convlab}, making it challenging to train the model effectively. Alternatively, \citet{wang-wong-2021-collaborative} proposed a multi-agent reinforcement learning framework where atomic actions were represented as the cross product of three spaces, with each part assigned to a different agent. While this approach showed promising results, it had the significant drawback of allowing only one atomic action per turn. This limitation necessitates more turns to achieve user goals and can lead to unnatural utterances. A recent work that used a transformer to generate actions is closely related to our research \cite{geishauser-etal-2022-dynamic}. However, their work primarily focused on continual learning and did not address the issue of action efficiency, which is a critical aspect of DPL. Our research aims to build upon these previous studies, addressing their limitations and enhancing the efficiency and effectiveness of DPL.

In this paper, we introduce a novel approach, the \textbf{Jo}int \textbf{T}ransformer \textbf{R}einforcement Learning Framework (\textbf{JoTR}), designed to address the limitations inherent in dialogue action candidates. The primary innovation of JoTR lies in its ability to generate actions with a word-level policy directly, a significant departure from the traditional reliance on fixed, human-defined templates. This is achieved by leveraging the capabilities of a text-to-text transformer, a model that has shown remarkable success in various natural language processing tasks. Simultaneously, JoTR employs reinforcement learning and reward shaping, two powerful techniques in machine learning, to optimize the word-level policy. This results in more efficient responses that require fewer interaction turns, a critical aspect in enhancing the user experience in dialogue systems. Furthermore, JoTR demonstrates improved learning efficiency compared to traditional methods, making it a promising approach for dialogue policy learning (DPL). We evaluate the performance of JoTR on two benchmark multi-domain dialogue modeling tasks, providing a comprehensive assessment of its effectiveness. Our research is primarily focused on developing a system that generates flexible and efficient responses for DPL, a critical aspect of dialogue systems. 

Our work offers several key contributions to the field. Firstly, we formulate the DPL as a conditional sequence generation problem, a novel approach that allows for more dynamic action generation. Secondly, we introduce a transformer reinforcement learning framework, designed to learn the word-level DPL and optimize it with a reward-shaping mechanism. This approach enhances the model's ability to adapt and improve over time. Thirdly, we validate the effectiveness of our proposed models through multi-domain testing on a simulator. This rigorous testing ensures that our model can handle a variety of dialogue scenarios effectively. Lastly, we implement a rule-based simulator for the benchmark Schema-Guided Dialogue (SGD) dataset \cite{rastogi2020towards}, a widely-used resource in dialogue system research. This implementation provides a practical tool for further research and development in the field.


\section{Related Work}
\paragraph{Dialogue Policy Learning}
The mainstream method for constructing a task-oriented dialogue (TOD) system is the pipeline approach, which divides the system into four interconnected modules: natural language understanding (to identify user intentions), dialogue state tracking (to monitor the dialogue status), dialogue policy learning (to decide the subsequent system action), and natural language generation (to generate dialogue responses) \citet{kwan2023survey}.
Reinforcement learning has been the main stream approach to optimize the dialogue policy \cite{levinLearningDialogueStrategies1997,singhReinforcementLearningSpoken2000,gasic-etal-2010-gaussian}.
To address the challenges of large state-action spaces and low exploration efficiency when applying reinforcement learning to DPL, hierarchical reinforcement learning has been used to break down the complex task into several subtasks and learn separate policies for these subtasks \cite{budzianowski-etal-2017-sub,peng-etal-2017-composite,kristianto-etal-2018-autonomous,tang-etal-2018-subgoal}. 
Alternatively, Other researchers have adopted reward learning and reward shaping to provide denser rewards and foster faster learning \cite{su-etal-2015-reward,su-etal-2016-line,wang-etal-2020-learning-efficient}.
Our approach aligns with this line of work and incorporates reward shaping to refine the dialogue policy.
Recently, a few studies have started to apply multi-agent reinforcement learning for dialogue policy learning \cite{liuIterativePolicyLearning2017,zhang-etal-2020-learning}. \citet{takanobu-etal-2020-multi} propose a joint learning process for the dialogue system and the user agent, and introduced the Hybrid Value Network to enhance the reward for each role. \citet{wang-wong-2021-collaborative} further expand on this approach by partitioning the action space into three subspaces and training a separate agent for each using multi-agent reinforcement learning. These works frame model DPL as a classification problem where the policy chooses a suitable dialogue action from a predefined action list. Contrasting to these methodologies, our approach obviates the need for a predefined action list, generating the dialogue action instead.

\paragraph{Pre-trained Language Model}
Recent studies have demonstrated significant advancements in TOD by fine-tuning pre-trained language models \cite{budzianowski-vulic-2019-hello,hosseini-asl_simple_2020,lee-2021-improving-end,su-etal-2022-multi,zhao-etal-2022-unids}. These studies have effectively unified the learning process of various modules within the pipeline approach. \citet{peng-etal-2021-soloist} introduces SOLOIST, a model that leverages GPT-2 \cite{radford2019language}, a pre-trained auto-regressive transformer decoder model. This model generates the belief state, dialogue action, and response in a sequential manner. 
SOLOIST is first fine-tuned on a large amount of dialogue corpus with a multi-task objective, then fine-tuned on the target dataset. \citet{he2022galaxy} further extends this approach by incorporating unlabelled dialogue data through semi-supervised learning. A notable drawback of sequential subtask generation is the error accumulation from earlier modules to the later modules in the pipeline. To mitigate this, PPTOD first formulates the belief state, then concurrently generates the dialogue action and response, thus preventing error propagation.  
These approaches require a large dialogue corpus to further pre-training and fine-tuning of the pre-trained language model. In contrast, our work only requires on small amount of data to warm up the model to obtain the state-of-the-art results. Additionally, our work utilizes reinforcement learning to fine-tune the model, which does not require labelled data. 

\section{JOTR}
\begin{figure*}[htbp]
\setlength{\belowcaptionskip}{-0.4cm}   
\centering
\includegraphics[width=1.8\columnwidth]{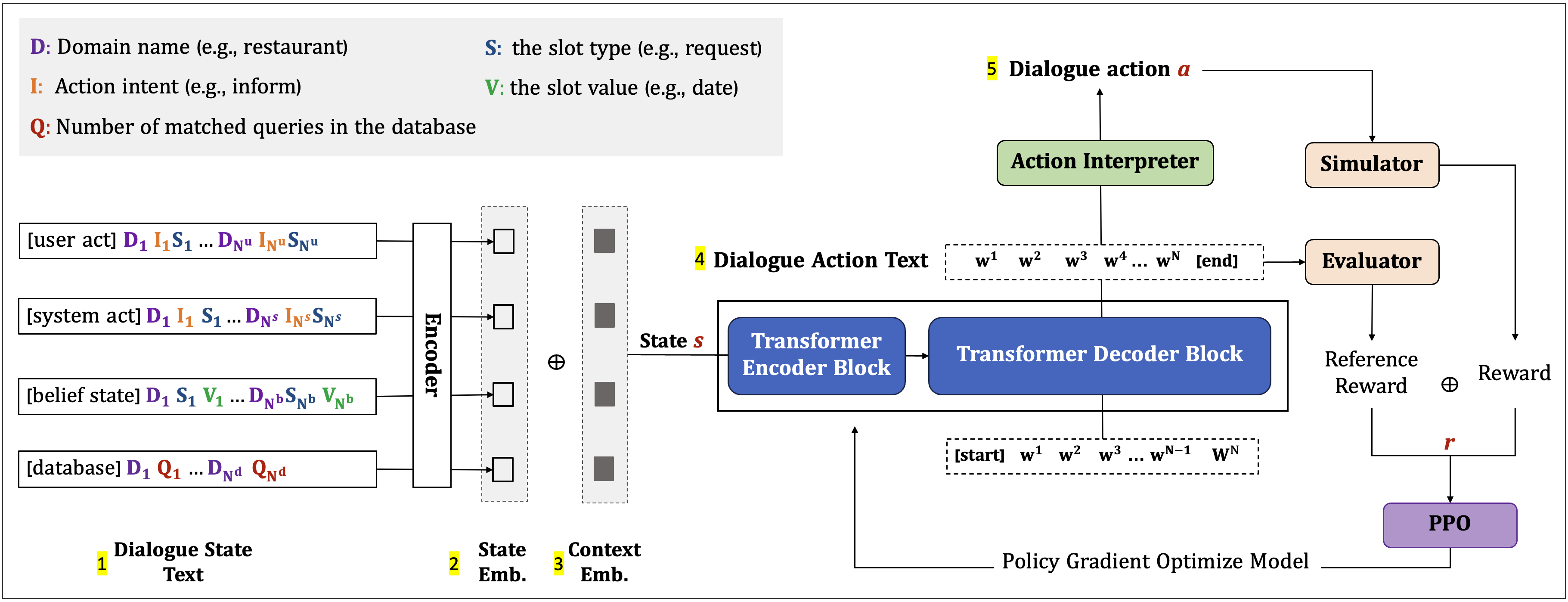}
\caption{The joint transformer and reinforcement learning framework illustration consists of: 1) (Left Part) Text Encoding - The encoder processes user act, system act, belief state, and database query results to form the state. 2) (Right Part) Model Optimization - The state directs action generation, with the Action Interpreter generating structured dialogue actions. The transformer-based policy model undergoes interactive optimization through reinforcement learning from scratch.}
\label{fig:arch}
\end{figure*}

We utilize a transformer-based model to directly generate dialogue actions, distinguishing it from conventional approaches that rely on selecting dialogue actions from a predefined set of atomic dialogue actions and their combinations. The architecture of our model is illustrated in Figure \ref{fig:arch}. In this section, We first provide a formal definition for DPL, then present the overview of our approach.

\subsection{Problem Definition}
The goal of DPL is to learn a policy that interacts with a user through the generation of dialogue actions, given the belief state and the database results, to satisfy the user goal $G=(C, R)$, where $C$ denotes the user constraints (e.g. an air ticket to Seattle) and $R$ represents the information required by the user (e.g. the price of the air ticket).  As mentioned in the previous section, the belief state keep tracks of the user's constraints throughout the dialogue. 
The belief state is defined as a list of domain, slot, value triplets (e.g. \textit{[(flight, destination, Seattle), (flight, day, tomorrow)]}).
The dialogue action is represented as a list of domain, intent, slot, value quadruples (e.g. \textit{[(flight, request, time, ?)]}).
An external database provides relevant entries to the dialogue policy based on the belief state. Figure \ref{fig1:intro} provides an illustrative example of such a process.

\subsection{Dialogue State Text Encoding}
The encoder generates state embeddings $e_u, e_s, e_b, e_d \in \mathbb{R}^d$ by encoding the flattened textual representations of four elements: user action, system action, belief state, and the database result. 
These linearized textual representations are referred to as the dialogue state text.
The user action is a sequence of tokens derived from atomic dialogue action triplets, each comprising the domain $D$, intent $I$, and slot $S$. This sequence is represented as $D_1, I_1, S_1, \ldots, D_{N^u}, I_{N^u}, S_{N^u}$, where $N^u$ denotes the number of atomic user dialogue actions. For instance, "\textit{hotel inform price hotel request name}" is a valid example.
The system action, similar to user action, is represented as $D_1, I_1, S_1, \ldots, D_{N^s}, I_{N^s}, S_{N^s}$, with $N^s$ indicating the number of atomic system dialogue actions.
The belief state is a sequence formed by concatenating the belief state triplets, where each triplet is composed of the domain $D$, slot $S$, and value $V$. This sequence can be represented as $D_1, S_1, V_1, \dots, D_{N^b}, S_{N^b}, V_{N^b}$. A typical example could be "\textit{hotel price expensive hotel area north}".
The database result is represented as $D_1, Q_1, \dots, D_{N^d}, Q_{N^d}$, with $N^d$ denoting for the number of queried domains and $Q$ denoting the number of matched entities in the database. An illustrative example would be “\textit{hotel 4 restaurant 2}”.

To obtain the state embeddings $e_u, e_s, e_b, e_d$, the [CLS] token, a common sentence representation placeholder, is prefixed to each dialogue state text \cite{devlin2018bert}. The output representation of the [CLS] token of each dialogue state text is used as the state embedding.
Initial experiments revealed inferior performance if the model is only fed with the state embeddings, likely due to the model's confusion about the varying types of state information being encoded.
Therefore, a context embedding was constructed for each dialogue state text. The context embeddings are added with the state embeddings individually to produce the state $s \in \mathbb{R}^{4 \times d}$.

\subsection{Word-level Dialogue Policy}
We have formulated the problem of dialogue policy learning (DPL) as a Markov Decision Process (MDP) on the word level. In this process, the system agent observes the current dialogue state $s$, executes an action $a$ (generated by the policy function), receives a response, a reward $r$, and the updated dialogue state $s'$. This cycle continues until the conversation ends. The action $a$ is textually represented as a sequence of words $w_{1:N} = w_{1}\ldots w_{N}$. The policy function can be detailed as a series of conditional probabilities:
\begin{equation}
\begin{aligned}
    \pi_{\theta}(a|s) = \prod_{i = 1}^{N} \mathcal{P}_{\theta}(w_i|w_{1:i-1}, s), 
    \label{e:2}
\end{aligned}
\end{equation}
where $\mathcal{P}$ is approximated with a transformer encoder-decoder network parameterized by $\theta$, representing the probability of the word $w_i$ condition on the preceding words and state. As shown in Figure \ref{fig:arch}, the transformer decoder generates the dialogue action text word by word, beginning with the start signal “[start]”, conditioned on the dialogue state, and proceeds until it encounters the stop signal “[end]”. 
Additionally, an action interpreter decodes the dialogue action text into structured format, populating slot values from the database, yielding the final dialogue action. 
This process involves verifying whether the dialogue action text adheres to the domain, intent, slot order, discarding any words that violate these conditions.
The policy $\pi$, parameterized by $\theta$, is optimized using reinforcement learning to minimize the negative expected cumulative future rewards: 
\begin{equation}
\begin{aligned}
    \mathcal{L}_{\theta} = -\mathbb{E}_{a_t \sim \pi_{\theta}(\cdot|s_t)}\left[\sum_{t=1}^T r(s_t, a_t)\right],
    \label{e:2}
\end{aligned}
\end{equation}
where $s_t$ and $a_t$ are the state and dialogue action turn $t$, and $T$ is the maximum turn. In practice, the expected gradient for a dialogue session can be approximated by using a Monte Carlo sample from $\mathcal{P}_{\theta}$. For each session example, the gradient is approximated as:

\begin{equation}
\begin{aligned}
    &\nabla_{\theta}\mathcal{L}_{\theta} \approx -\sum_{t=1}^T r(s_t, a_t)\nabla_{\theta}\log \pi_{\theta}(a_t|s_t)
    = \\&-\sum_{t=1}^T r(s_t, w^t_{1:N^t}) \sum_{i=1}^{N^t} \nabla_{\theta} \log \mathcal{P}_{\theta}(w_i^t|w_{1:i-1}^t, s_t), 
    \label{e:3}
\end{aligned}
\end{equation}
where $N^t$ is the length of the action text at turn $t$. 

\subsection{JoTR for Efficient Policy Training}
We employ Proximal Policy Optimization (PPO) \cite{schulman2017proximal} to optimize the policy. More specifically, we minimize the objective function for each session example.

\begin{equation}
\begin{aligned}
    \mathcal{L}_\theta 
    &= \sum_{t=1}^T -\hat{\mathbb{E}}_{t}\left[\min \left(\frac{\pi_{\theta}\left(a_{t} \mid s_{t}\right)}{\pi_{\theta_{\text {old }}}\left(a_{t} \mid s_{t}\right)}  \hat{A}_{t}^\phi, \right . \right .\\ & \left . \left . \operatorname{clip}\left(\frac{\pi_{\theta}\left(a_{t} \mid s_{t}\right)}{\pi_{\theta_{\text {old }}}\left(a_{t} \mid s_{t}\right)}, 1-\epsilon, 1+\epsilon\right) 
    \hat{A}_{t}^\phi \right)\right]  \\
    &= \sum_{t=1}^T -\hat{\mathbb{E}}_{t}\left[\min \left(\frac{    \prod_{i = 1}^{N}\mathcal{P}_{\theta}(w_i|w_{1:i-1}, s_t)}{\prod_{i = 1}^{N}\mathcal{P}_{\theta_{\text{old}}}(w_i|w_{1:i-1}, s_t)}  \hat{A}_{t}^\phi, \right . \right .\\ & \left . \left . \operatorname{clip}\left(\frac{\prod_{i = 1}^{N}\mathcal{P}_{\theta}(w_i|w_{1:i-1}, s_t)}{\prod_{i = 1}^{N}\mathcal{P}_{\theta_{\text{old}}}(w_i|w_{1:i-1}, s_t)}, 1-\epsilon, 1+\epsilon\right) \hat{A}_{t}^\phi\right)\right], 
\end{aligned}
\end{equation}
where $\hat{A}_{t}^\phi = r(s_t, a_t) + \gamma V^\phi(s_{t+1}) - V^\phi(s_t)$ is the advantage estimation and $V^\phi(s_t)$ is the value function estimated by the critic parameterized by $\phi$.





To improve the efficiency and quality of the dialog response, we integrate reward-shaping into the reinforcement learning fine-tuning process. 
The goal is to prevent the policy from generating protracted yet predominantly irrelevant dialogue actions during optimization with PPO. 
To achieve this, we propose a reward-shaping function assigning supplementary rewards to guide the model to learn.
Formally, we replace the $r(s_t, a_t)$ with $\hat{r}(s_t,a_t, G)$ defined as
\begin{equation}
\begin{aligned}
    \hat{r}(s_t,w_{1:N^t}^t, G) = r(s_t, a_t) + F(G, w_{1:N^t}^t),
\end{aligned}
\end{equation}
where $G$ denotes the user goal and $F$ represents the shaping reward. We design $F$ to provide different reward based on the follow: 
(1) If the system informs a slot present in the user's request slot list, it receives an additional $\lambda$ reward. Conversely, informing other slots receives an additional -1 reward. (2) If the system requests a slot included in the user's inform slot list, it receives an additional $\lambda$ reward. However, requesting other slots results in an additional -1 reward. $\lambda$ is a hyperparameter that controls the aggressiveness of the dialogue agent to inform or request additional slots. We try $\lambda$ with value 3, 4, 5, 6, 7. We find that the range of 3 to 5 yielded favourable results during validation. 
Higher $\lambda$ values encourage the dialogue agent to attempt many actions in a single turn, where one successful inform or request action offsets the negative rewards incurred by other irrelevant actions. We pick $\lambda=3$ for all the experiments.

\section{Experiments and Results}
Experiments are carried out on MultiWOZ 2.0 \cite{budzianowski2018multiwoz}, utilizing a publicly accessible agenda-based user simulator \cite{zhu2020convlab}, and on the SGD dataset with our developed rule-based simulator. Furthermore, we incorporate human evaluations, in which evaluators interact with various models and assess the success of the dialogue upon its completion. While all models, except SimpleTOD, are optimized in the dialogue action space, SimpleTOD takes the utterance dialogue history as input and generates both the dialogue action and the system utterance.

\section{Evaluation Metrics}
\label{sec:appendix_simulator}
We focus on three evaluation metrics to align with previous work \cite{wang-wong-2021-collaborative}: success rate, average number of turns, and average reward. A dialogue session is considered successful if it fulfils all the user requests, and reserves an entity that meets the user's specifications if necessary. 
The average number of turns is calculated by counting the number of interactions between the two parties, with each full interaction counted as two turns. 
The average reward is the total cumulative reward obtained in each dialogue session. 
Additionally, since SimpleTOD uses dialogue utterances as input, an NLG component is needed to convert the user's dialogue actions into utterances. However, the performance of SimpleTOD is significantly dependent on the NLG component. Therefore, we also evaluate it against a testing corpus, which we believe is a superior method than using a user simulator equipped with NLU and NLG modules that could potentially introduce noise.

\begin{figure*}
\small
\centering
\includegraphics[width=0.45\textwidth]{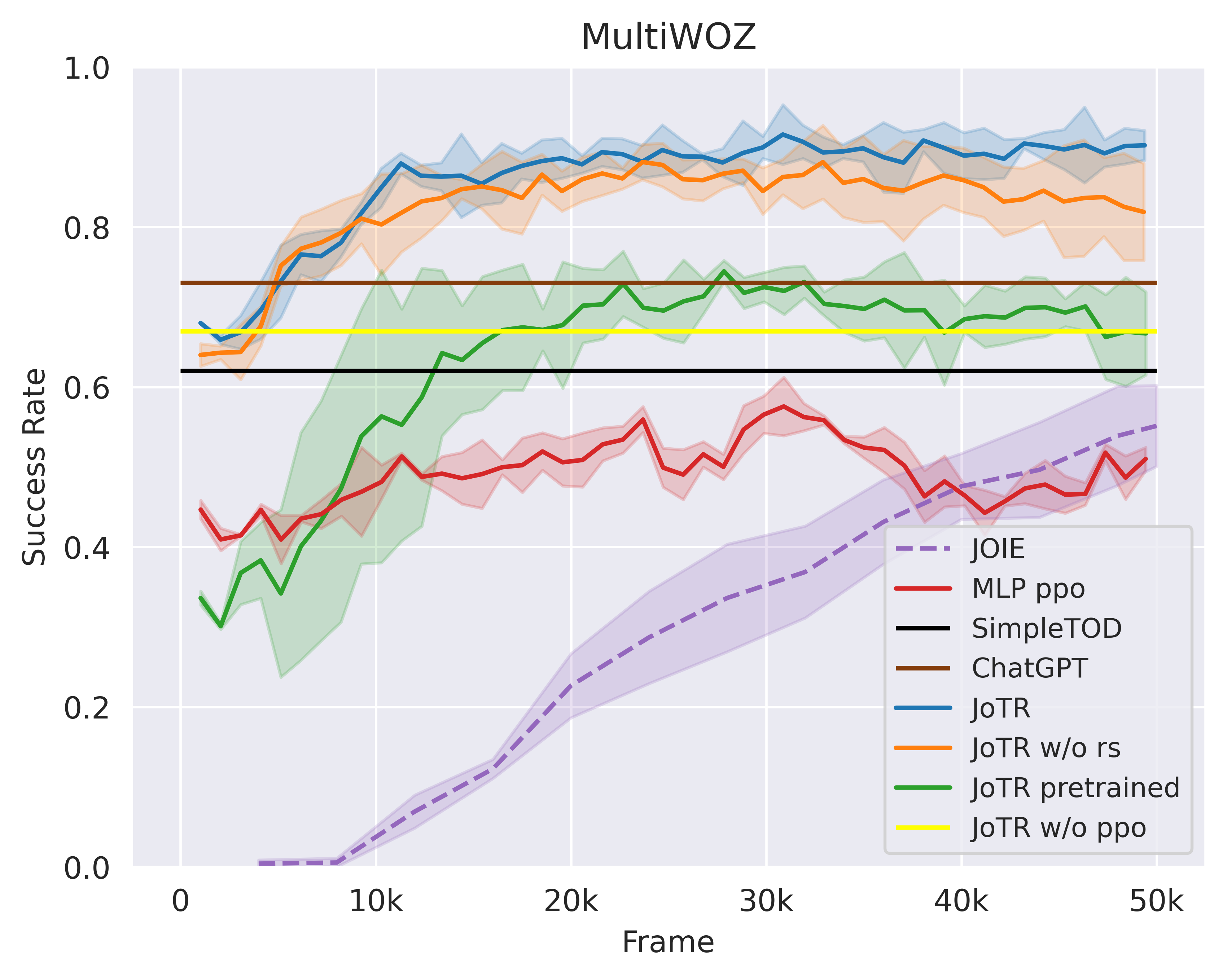}
\centering
\includegraphics[width=0.45\textwidth]{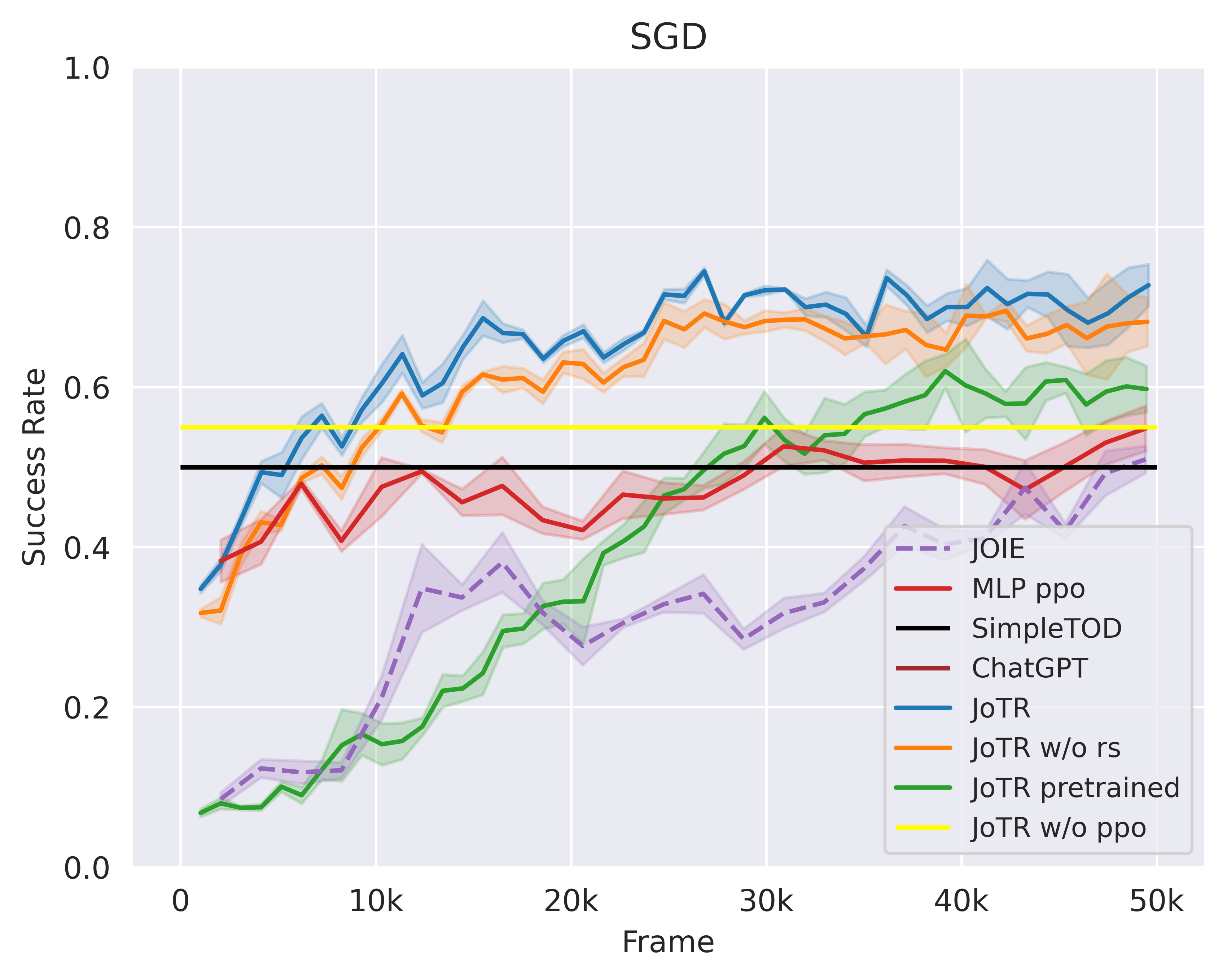}
\caption{The learning curve of various models on MultiWOZ and SGD, with the mean and standard deviation illustrated over 5 runs. We didn't reach JOIE's 400k frames, as JoTR attains a 0.93 success rate with only 50k frames, surpassing JOIE, and training 400k frames for its variants is costly.}
\label{fig:learning_curve}
\vspace{-3mm}
\end{figure*}

\section{Training and Implementation Details}
\label{sec:implementation}


We implement all the models in PyTorch \cite{paszke2019pytorch} and Transformers \cite{wolf-etal-2020-transformers}. 
We use a randomly initialized transformer encoder with 1 hidden layer, 1 head, and hidden size of 256 as the dialogue state text encoder of JoTR. 
In JoTR$_\text{pretrained}$, we use DistilBert \cite{sanh2019distilbert} with pretrained weights from Huggingface as the initial weights for the dialogue state text encoder.
For the encoder-decoder model, we use a transformer with 1 hidden layer, 1 attention head, and hidden size of 256 for both the encoder and decoder. The total number of parameters is 5M. All variants of JoTR and MLP$_\text{ppo}$ are warmed up before reinforcement learning by first performing supervised learning on a training set to be consistent with previous work \cite{wang-wong-2021-collaborative}. We use the same set of 10K dialogue turns sampled randomly from the original training set to warm up all models. A separate, non-overlapping set of 3K dialogue turns is used for validation in the warm-up phase.  The same set of hyperparameters are used on MultiwWOZ and SGD in both pretraining and PPO fine-tuning. For the warm-up phase, the models are trained using a batch size of 32 and a learning rate of $3 \times 10^{-4}$. The models are trained for 80 epochs but we include an early stopping mechanism that halted training when no improvement is observed in the validation set over five consecutive epochs. In PPO training, we use an actor learning rate of $5 \times 10^{-7}$ and critic learning rate of $1 \times 10^{-4}$. The critic is a transformer with identical architecture as the encoder-decoder model.
The models are trained with a total of 50K frames. 
All models are trained on a single RTX 3090 GPU, which take about 10 hours to train.
The maximum interaction turn allowed is 40. The main reward provided by the environment is -1 in every turn, and a reward of 80 or -40 at the end for successful or failed dialogue respectively. 

The prompt for ChatGPT is shown in Table 1. Following previous work \cite{wang-etal-2022-super}, we provide one formatting example in the zero-shot setting. In our preliminary experiments, we find that the task definition and the output specification significantly influence ChatGPT's ability to understand the task and generate a structured output. 

\begin{table}[htbp]
    \centering
    \tiny
    \begin{tabular}{p{.95\linewidth}}
    \toprule
     \textbf{Prompt:}
     \\
     \textcolor{mypurple1}{You are a dialogue agent to assist me with my queries and provide me with relevant information from a database. My questions are formatted as tuples of (domain, intent, slot, slot value) accompanied by the number of matching results that satisfy my constraint from the database, e.g., "4 matches".} 
    \\ \textcolor{myyellow1}{Your responses should be formatted as one or several tuples of (domain, intent, slot) to provide me with the necessary information.} 
    \\ \textcolor{myyellow1}{The domain is selected from attraction, hospital, ....} 
    \\ \textcolor{myyellow1}{The intent is selected from inform, request, ....} 
    \\ \textcolor{myyellow1}{The slot includes addr: the address of the hotel, attraction, restaurant, hospital, or police station....}   
    \\ \textcolor{myblue1}{Example 1:} 
    \\ \textcolor{myblue1}{USER: [(train, inform, depart, london kings cross)] 3 matches.} 
    \\ \textcolor{myblue1}{ASSISTANT: [(train, inform, id)]} 
    \\ ... 
    \\ \textcolor{myred1}{Example 2:} 
    \\ \textcolor{myred1}{USER: [(restaurant, inform, area, centre)]}... 
    \\ \hline
       \textbf{Output:}  \\
       ASSISTANT: [(restaurant, request, day), (restaurant, request, time), (restaurant, request, people)]\\
    \bottomrule
    \end{tabular}
    \caption{A zero-shot ChatGPT prompt example for dialogue act prediction. It consists of \textcolor{mypurple1}{task definition}, \textcolor{myyellow1}{output specification}, \textcolor{myblue1}{formatting example} and \textcolor{myred1}{dialogue history}. }
    \label{tab:chatgpt_prompt}
\end{table}

\section{Simulator Details}
We implement an agenda-based simulator for SGD, following the prevalent approach for user simulator design as outlined in previous works \cite{schatzmann_agenda-based_2007,wang-wong-2021-collaborative, kwan2023survey}. The agenda-based simulator samples a user goal according to the distribution of slots in the training set of SGD. The user goal is kept unknown to the dialogue agent.
It maintains a stack (i.e. user agenda) that stores all the user actions that the user needs to inform perform to achieve his/her goal during the conversation. It acts to the system's actions according to some pre-fined rules.

\subsection{Baseline Agents}
We compare our model, JoTR, to five other models. 1) \textbf{JOIE} \citep{wang-wong-2021-collaborative}, the current state-of-art(SOTA) on MultiWOZ, is a collaborative multi-agent model that generates a single dialogue action without a predefined action list. 2) \textbf{MLP$_{\text{ppo}}$}, an agent optimized with PPO with fixed action candidates. 3) \textbf{SimpleTOD} \citep{hosseini-asl_simple_2020}, a GPT-2-based agent trained with supervised learning to generate dialogue actions along with belief states and responses based on the dialogue history. 4) \textbf{DASP} \citep{jhunjhunwala-etal-2020-multi}, an LSTM-based agent that trained with human supervision to select among N-best action candidates based on the dialogue history. 5) \textbf{ChatGPT} is a large language model developed on the foundation of InstructGPT \cite{ouyang2022training}. It functions as an efficient conversational agent, capable of interpreting user prompts and producing logically consistent replies. ChatGPT has demonstrated impressive results across a diverse range of natural language processing assignments. We use it to generate dialogue actions based on the dialogue action history using a zero-shot prompt as shown in Table \ref{tab:chatgpt_prompt}. 

\begin{figure*}[htbp]
\centering
\includegraphics[width=0.99\textwidth]{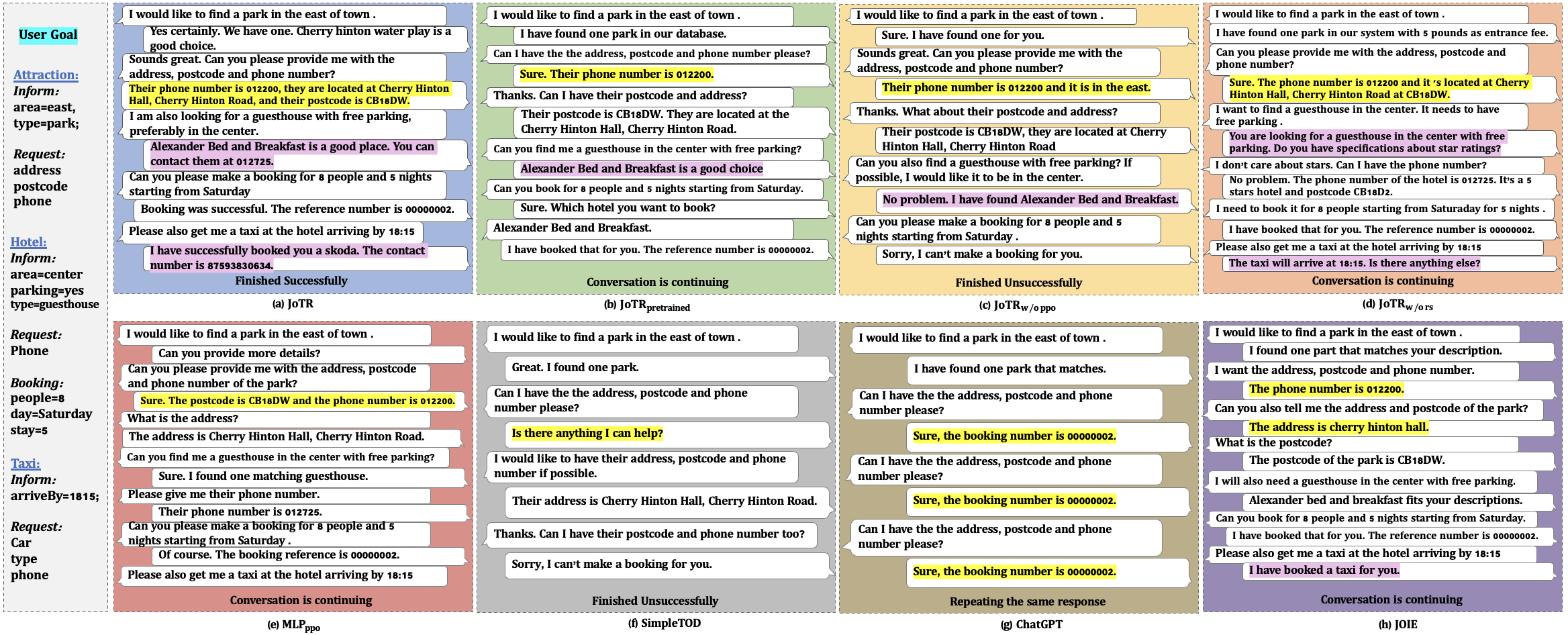}
\caption{An illustrative dialogue example featuring responses from various models is provided. The system's dialogue actions (text on the right) highlighted in \textcolor{yellow}{yellow} emphasize JoTR's capability to manage complex and out-of-domain user actions, which other models struggle to handle. The dialogue actions in \textcolor{pink}{pink} showcase JoTR's proficiency in preemptively informing relevant slots, an area where other models fall short.
\vspace{-0.5em} and}
\label{fig:case}
\end{figure*} 

To further demonstrate the advantages of our model, JoTR, we also compare it to three variants. \textbf{JoTR$_{\text{w/o rs}}$} does not use reward shaping. \textbf{JoTR$_{\text{w/o ppo}}$} is only pretrained with supervised learning and not further fine-tuned with PPO. \textbf{JoTR$_{\text{pretrained}}$} uses a pre-trained BERT as the context encoder but with the weights fixed.


\subsection{Main Results}
\begin{table}[]
    \centering
    \renewcommand{\arraystretch}{0.9}
    \small
    \setlength{\tabcolsep}{0.4mm}{
    \begin{tabular}{l lll lll }
        \toprule
         \multirow{2}{*}{Model} &  \multicolumn{3}{c}{MultiWOZ} & \multicolumn{3}{c}{SGD} \\
         \cmidrule(lr){2-4}\cmidrule(lr){5-7}& Succ.$\uparrow$ & Turn$\downarrow$ & Rew.$\uparrow$ & Succ.$\uparrow$ & Turn$\downarrow$ & Rew.$\uparrow$ \\
         \midrule 
         JOIE$\dagger$ & 0.91 & 18.90$^*$ & 40.82 &0.51 &11.10$^{*}$ &15.32 \\
         MLP$_{\text{ppo}}$ & 0.56 & 30.72 & -26.76 & 0.54 & 23.43 & 16.50 \\
         SimpleTOD$\ddagger$ & 0.62 & - & - & 0.50 & - & - \\
         DASP$\dagger,\ddagger$ & 0.85 & - & - &0.70 & - & - \\
         ChatGPT & 0.73 & 13.10 & 41.05 & 0.50 & \textbf{11.04} & 15.48 \\
         \midrule 
         JoTR & \textbf{0.93} & \textbf{9.94} & \textbf{68.46} & \textbf{0.79} & 15.23 & \textbf{49.25} \\
         JoTR$_{\text{w/o rs}}$ & 0.89 & 9.95 & 66.42 & 0.72 & 16.53 & 38.84 \\ 
         JoTR$_{\text{w/o ppo}}$ & 0.67 & 18.44 & 32.18 & 0.55 & 24.76 & 14.62 \\
         JoTR$_{\text{pretrained}}$ &0.76 & 14.19 & 44.87 & 0.64 & 19.25 & 28.18 \\
         \bottomrule
    \end{tabular}
    \caption{
    Dialogue act modeling performance is measured by success rate (Succ.), turn, and reward (Rew.) All the RL-based agents use the same reward values and assignments during testing. Results with $\dagger$ are from original papers. We reproduce JOIE to obtain the performance in SGD. Models with $\ddagger$ are tested against a corpus. The best results are in bold. *In our settings, a user-system utterance pair counts as 2 turns, while in the JOIE paper, it's 1 turn.
    }
    \vspace{-3mm}
    \label{tab:results}
    }
\end{table}

\footnotetext{https://github.com/salesforce/simpletod}
The learning process demonstrated in Figure \ref{fig:learning_curve} showcases the superior performance and efficiency of the proposed JoTR model. 
Notably, JoTR outperforms the previous SOTA model JOIE (0.93 vs 0.91 for MultiWOZ and 0.79 vs 0.51 for SGD) despite only trained with 50K frames, compared to JOIE's 400K frames \cite{wang-wong-2021-collaborative}. Moreover, the ability of JoTR to improve significantly with such short training makes it more suitable for real-world applications. Note that we implement JOIE and obtain the result in SGD with 50K frames while the result in MultiWOZ is from the original paper. 

Table \ref{tab:results} shows that JoTR requires significantly fewer turns than JOIE to satisfy the user goal. 
This efficiency can be attributed to JoTR's capacity to generate multiple atomic actions in one turn, in contrast to JOIE's single action prediction. This characteristic not only reduces the total number of interaction turns but also enhances JoTR's practicality for everyday use. All JoTR variants outperform MLP$_{\text{ppo}}$. Notably, JoTR$_\text{w/o ppo}$ surpasses MLP$_{\text{ppo}}$ without additional RL fine-tuning, indicating that the strong learning capacity of transformer is effectively harnessed for learning the specific structural properties of dialogue actions. In comparison to SimpleTOD, JoTR performs better due to its use of dialogue actions as input, which reduces noise and complexity compared to SimpleTOD's language utterances. 
Furthermore, JoTR's encoder-decoder model structure aids in capturing context information more effectively than SimpleTOD's decoder-only model. Lastly, JoTR performs significantly better than ChatGPT. Most errors made by ChatGPT can be categorized into two categories: 1. hallucination on domain, slot and values. 2. Violations of output format constraints. 

\subsection{Ablation Study}
\subsubsection{The Effectiveness of Reward Shaping}
The success rate increase from 0.89 to 0.93 in MultiWoz and 0.72 to 0.79 in SGD when reward shaping is applied. It manifests that reward shaping plays a big role in achieving high success rate. Furthermore, Figure \ref{fig:learning_curve} demonstrates that JoTR consistently maintains a higher success rate throughout the fine-tuning process.  
This observation highlights the advantage of using a dense and well-designed reward in RL fine-tuning, corroborating previous findings \cite{wang2022integrating}.

\subsubsection{The Necessity of RL Fine-Tuning}  JoTR$_\text{w/o ppo}$ underperforms significantly without continuously being optimized with RL, exhibiting a reduction of as much as 28\% in the success rate on MultiWOZ. This underlines the critical role of RL fine-tuning in refining the behavior of the policy model through the reward.

\subsubsection{The Importance of Training from Scratch} JoTR significantly outperforms JoTR$_\text{pretrained}$ (0.93 vs 0.76 in Multiwoz, 0.79 vs 0.64 in SGD). 
As evidenced by Figure \ref{fig:learning_curve}, JoTR$_\text{pretrained}$ exhibits a notably lower success rate initially. This can be attributed to the distinct structure of the dialogue actions' input space, which markedly differs from the natural language space where the model was originally pretrained. 
This discrepancy cannot be bridged effectively by supervised training during the warm-up phase or reinforcement learning in the fine-tuning phase.

\subsection{Case Study} \label{sec:case_study}

\begin{figure}
    \centering
    \includegraphics[width=0.47\textwidth]{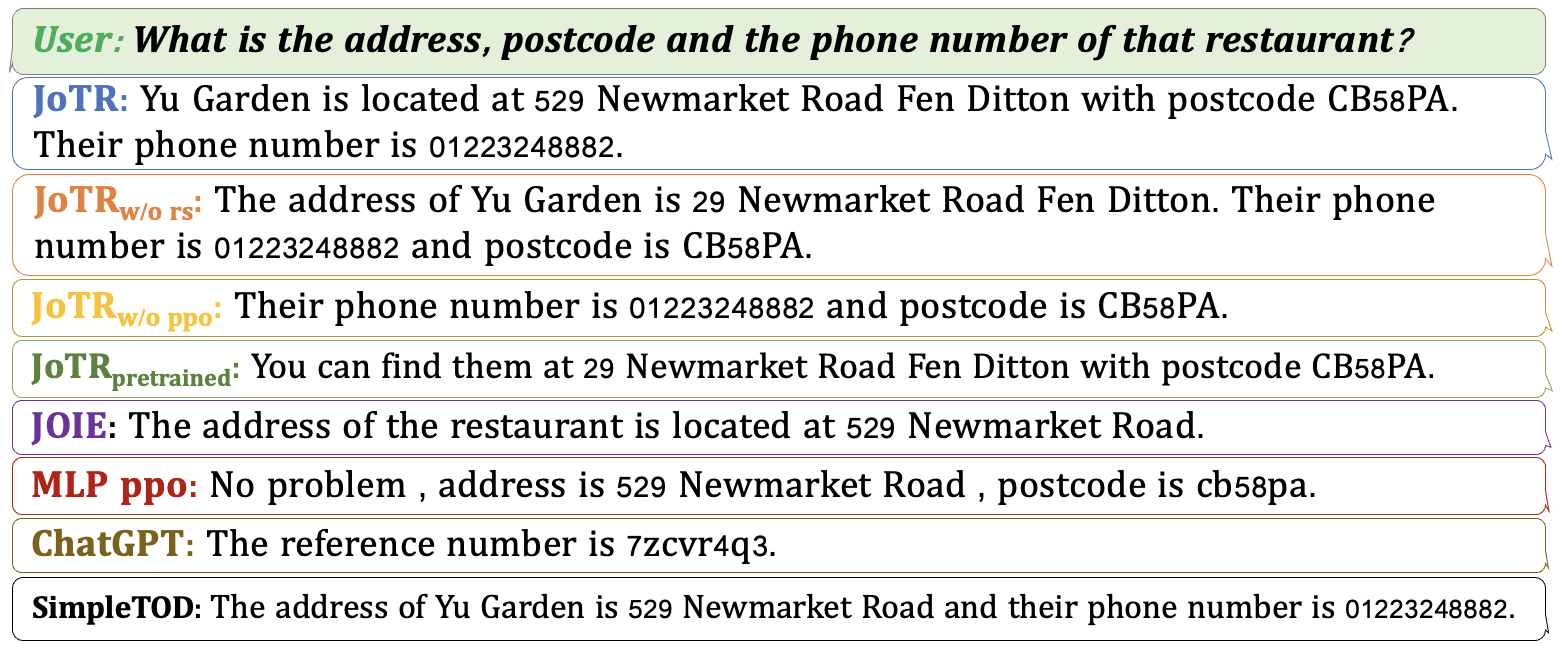}
    \caption{An example user utterance accompanied by the language responses corresponding to the dialogue actions predicted by various agents. We used the rule-based NLG component in ConvLab2\footnotemark to generate the language response given the predicted dialogue action.}
    \vspace{-1.5em}
    \label{fig:case_extract}
    
\end{figure}

As demonstrated in Figure \ref{fig:case_extract}, we observe that when the user requests a slot combination that is never seen in the training set, both JoTR and JoTR$_\text{w/o rs}$ are able to successfully informed all requested slots. This demonstrates their robust ability to generate effective and efficient dialogue actions. 
In contrast, JoTR$_\text{w/o ppo}$, JoTR$_\text{w/o pretrained}$, and SimpleTOD were unable to inform all requested slots, potentially a reflection of their inferior performance relative to JoTR.
JOIE only informed one requested slot, likely due to its design limitation of generating a single action per turn. 
Moreover, MLP$_{\text{ppo}}$ could not inform the complete slots as well, since the dialogue action for informing address, postcode, and phone number is not found within its predefined action set.
Lastly, ChatGPT responded inappropriately, for reasons elaborated in the preceding section.  

We also provide a full dialogue example of various models interacting with the user simulator in Figure \ref{fig:case}. The user requested for the address, postcode and phone of the park in the second turn. There is not a single training example that request these three slots simultaneously, which serves to showcase ability of different models to respond with complex and out-of-domain user actions. In consistent with Figure \ref{fig:case_extract}, JoTR and JoTR$_{\text{w/o rs}}$ were able to inform all three slots while other models can't (highlighted in yellow) Furthermore, when the user requested for a guesthouse in turn three, JoTR is able to provide the phone number without being explicitly asked while JoTR$_{\text{w/o rs}}$ and other models failed to do so as (highlighted in pink). This illustrates the rewarding shaping can incentivize the model to provide additional information preemptively. In this example, we can also see the dialogues of other models are significantly longer than those of JoTR. Therefore, JoTR is able to achieve the user's goal efficiently.

\subsection{Human Evaluation}
\label{sec:human-eval}
\begin{table}[htbp] 
\vspace{-3mm}
\setlength{\belowcaptionskip}{-0.4cm}   
\centering
\small
\setlength{\tabcolsep}{1.8mm}{
\begin{tabular}{l c c}
\toprule 
Model&Succ.(MultiWOZ)$\uparrow$ &Succ.(SGD)$\uparrow$ \\
\midrule 
JOIE                        &0.56 &0.53   \\
MLP$_{\text{ppo}}$          &0.52 &0.56 \\
SimpleTOD                   &0.62 &0.50 \\
DASP                        & -   & -  \\
ChatGPT                     &0.66 &0.52 \\
\midrule 
JoTR                        &\textbf{0.92} &\textbf{0.76}\\
JoTR$_{\text{w/o rs}}$      &0.84 &0.70\\
JoTR$_{\text{w/o ppo}}$     &0.66 &0.56\\
JoTR$_{\text{pretrained}}$  &0.68 &0.60\\
\bottomrule
\end{tabular}}
\vspace{-0.5em}
\caption{\label{tab:human-evaluation} Human evaluation results. We use the models trained with 50K frames for all agents. }
\end{table}

\footnotetext{https://github.com/thu-coai/ConvLab-2}
We further conduct a human evaluation to validate the simulation results using the models trained with 50K frames. 
We recruit 3 volunteer student helpers as evaluators to interact with different models.  
For each model, we held 50 dialogue sessions.
In each session, an evaluator is assigned a randomly selected model and user goal.
The evaluators are instructed to interact with the model in accordance with the user goal, with a maximum of 20 turns per session, aligning with the settings used in the experiments in previous sections.
At the end of each session, the evaluators assessed the success or failure of the dialogue. 
The results are illustrated in Table \ref{tab:human-evaluation}, which are consistent with the previous results using a user simulator.

\section{Conclusion}
We introduced JoTR, a versatile framework for dialogue policy learning using joint text-to-text transformer reinforcement learning. It trains word-level policies that can generate dialogue actions without the need for predefined templates. Empirical results from two benchmark datasets show that our model, which does not rely on predefined action templates, outperforms the strongest baseline in terms of both policy learning efficiency and dialogue action quality as determined by simulated and human evaluations.


\bigskip

\bibliography{aaai24,anthology}

\end{document}